\documentclass[letterpaper, 10 pt, conference]{ieeeconf}  

\IEEEoverridecommandlockouts                              

\usepackage{graphicx}
\usepackage{epstopdf}
\usepackage{multicol}
\usepackage{stfloats}
\usepackage{amsfonts}

\usepackage{mathrsfs}
\usepackage{multicol}
\usepackage{stfloats}
\usepackage{cite}
\usepackage{amsfonts}
\usepackage{mathrsfs}
\usepackage{amsfonts}
\usepackage{empheq}
\usepackage[linesnumbered,ruled,lined,boxed]{algorithm2e}
\usepackage[procnames]{listings}
\usepackage{algpseudocode}
\usepackage{color}
\usepackage[ruled]{algorithm2e}
\usepackage{fancybox}
\usepackage{framed}
\usepackage{amsfonts, color}
\usepackage{color}
\usepackage{setspace}
\usepackage[dvips]{epsfig}
\usepackage{framed}
\usepackage{psfrag, amssymb, amsmath, cite}
\usepackage[colorlinks,linkcolor=red,anchorcolor=blue,citecolor=blue]{hyperref}
\DontPrintSemicolon

\usepackage{verbatim}
\usepackage{graphicx, subfigure}

\usepackage{url}

\title{\LARGE \bf
Deep Robotic Grasping Prediction with Hierarchical RGB-D Fusion
}

\author{Yaoxian Song$^{1}$$^{2}$, Jun Wen$^{3}$, Yuejiao Fei$^{1}$, Changbin Yu$^{1}$
	\thanks{$^{1}$The authors are with School of Engineering at Westake University Emails: 
		{\tt\small \{songyaoxian, feiyuejiao, yuchangbin\} @westlake.edu.cn}}%
	\thanks{$^{2}$The author is also PhD students in Computer Science at Fudan University.}
	\thanks{$^{3}$The author is with College of Computer Science and Technology at Zhejiang University Email:
		{\tt\small junwen@zju.edu.cn}}
}

\begin{document}

\maketitle
\thispagestyle{empty}
\pagestyle{empty}

\begin{abstract}
Robotic arm grasping is a fundamental operation in robotic control task goals. Most current methods for robotic grasping focus on RGB-D policy in the table surface scenario or 3D point cloud analysis and inference in the 3D space. Comparing to these methods, we propose a novel real-time multimodal hierarchical encoder-decoder neural network that fuses RGB and depth data to realize robotic humanoid grasping in 3D space with only partial observation. The quantification of raw depth data's uncertainty and depth estimation fusing RGB is considered. We develop a general labeling method to label ground-truth on common RGB-D datasets. We evaluate the effectiveness and performance of our method on a physical robot setup and our method achieves over 90\% success rate in both table surface and 3D space scenarios. The video is available in \url{https://youtu.be/_iRyLcfbTfg}.
\end{abstract}

\section{Introduction}
Robotic grasping addresses the control of a robot to grasp and transfer. Rapid and reliable robotic grasping has been researched for decades and is useful for many applications, e.g., manufacturing, retailing, and service industry\cite{bicchi2000robotic}. For most of them, being able to grasp and transfer objects in unstructured environments is critical, which is receiving increasing attention from the robot community.   

The key to successful robotic grasping lies in the modeling of a mapping from the perception space to the action space. Traditional methods usually rely on the precise estimation of shape and pose of objects\cite{mason1985mechanics}\cite{lozano1986motion}. To achieve precise grasping, they need to aggregate together object geometric knowledge, environment configuration and robot setup information to design a controller, which actually can hardly be precisely obtained in real applications. In most scenarios, object geometric is merely partially observable, and the environments are too complicated to model directly. To relieve these problems caused by uncertainty, \cite{schwarz2017nimbro} \cite{zeng2017multi} propose to split the grasping task into several steps with the Amazon Picking Challenge. They first do object detection or segmentation from the observation and then fit a 3D model to a segmented point cloud to recover the 6D pose of objects to generate final arm motions. Despite the improved performance, they are too computationally expensive, which limit their applications, especially those calling for real-time implementation. Besides, the task splitting can cause additional system uncertainty, which harms control reliability.


Another approach for robotic grasping is based on end-to-end learning, directly mapping observations to grasping policies. \cite{mahler2017dex} trains a deep neural network to generate grasping policy from synthetic datasets. In the same line, instead of generating grasping policy by time-consuming candidates sampling, \cite{morrison2018closing} predicts the grasping action using an encoder-decoder based neural network with depth images in real-time. In these methods, only depth features are exploited, without incorporating RGB features, which may cause them suffer from observation noise, viewpoint variations, etc.



%

In this paper, we explore a novel method for robust robotic grasping control in unstructured environment based on multimodal data fusing.  We hierarchically fuse RGB features and depth features using a novel deep architecture, named UG-Net V3, to learn the mapping from partial observations to the grasping space. Jointly incorporating RGB and depth features on the eye-in-hand camera system enables reducing observation noise and environmental uncertainty and thus improve control robustness. Our main contribution can be summarized as follows:
\begin{itemize}
    \item We propose a novel deep network to fuse RGB features and depth features hierarchically to generate grasping policy from eye-in-hand camera systems. We achieve humanoid grasping end-to-end without Iterative closest point(ICP) or object reconstruction.
	\item Instead of directly obtaining from inpainted depth images, we quantify the uncertainty of raw depth images and reconstruct depth images by RGB-D fusion.
	\item We experiment with an open-loop controller for robotic grasping in both table surface and 3D scenarios. Comparing to the state-of-the-art, we achieve competitive performances in complicated and various scenarios. As a byproduct, a novel label method is proposed, enabling to train a grasping neural network without human-labeled ground-truth by exploiting easily available computer vision datasets with RGB, depth, and object mask images contained.
\end{itemize}

\section{Related work}
\subsection{Learning based robotic grasping}

Traditional analytic methods concentrate on creating a 3D object models database and register sensor data to known object in the database\cite{murray2017mathematical}\cite{siciliano2016springer}\cite{kehoe2013cloud}\cite{weisz2012pose}. These methods do not generalize well in a large variety of novel objects in practice\cite{bohg2013data}. Learning-based methods become popular with the widely adoption of deep learning technic. \cite{detry2013learning}\cite{lenz2015deep}\cite{kappler2015leveraging} \cite{morrison2018closing} use deep learning to generate grasping policy. \cite{zeng2018learning}\cite{thomas2018learning}\cite{balasubramanian2012physical}\cite{levine2018learning}\cite{sadeghi2018sim2real} design controller using deep reinforcement learning by exploring   robotic task environment and finally realize robotic grasping.

\subsection{Multimodal perception}
Robotic control with multisensor data fusion has been researched for many years in the control field. Robotic sensors usually consist of RGB-D camera, inertial measurement unit, laser radar, haptic device. The complementary nature of different sensor modalities have been explored in robotic inference and decision making. \cite{zeng2019deep} adopts RGB and depth image to estimate the surface normal of the object. \cite{mahler2017dex}\cite{lenz2015deep}\cite{arruda2016active}\cite{Pas2017Grasp}\cite{sung2017learning}\cite{liu2017learning} based on RGB image with depth or point cloud to predict grasping policy. \cite{bekiroglu2011learning} \cite{calandra2018more}\cite{gao2016deep}\cite{sinapov2014learning} fuses RGB and haptic data to train a grasping network. In \cite{lee2019making}, not only visual and force-torque data is used but also the robotic proprioception, including position and velocity of the end-effector, is also used to predict grasp representation. 


Most methods mentioned above fuse more than one modality data to represent the observation on a single scale. We can still tap the potential of multiple scales following \cite{zeng2019deep}. Therefore, we propose our  hierarchical encoder-decoder neural network to complete multi-task prediction shown in Fig \ref{UG-Net_V3}.

\section{Preliminary and problem formulation} \label{Sec:preliminary}
    
\subsection{Grasp definition}\label{Sec:grasp_define}
\textbf{Grasp space representation} We propose an 8D grasp representation \eqref{eq:grasp_def} including 6D object pose $P=(x,y,z, \gamma_x, \beta_y, \alpha_z)$, gripper's orientation and opening width which is similar with \cite{morrison2018closing}\cite{jiang2011efficient}. Considering we only utilize RGB-D data from a single point of view, the object is partially observable and hard to achieve its precise pose estimation without sufficient prior knowledge. Therefore, we estimate grasp representation except 3D pose from RGB-D fusion algorithm and obtain 3D pose from the proprioception of the eye-in-hand system mentioned in Sec. \ref{sub_3d_test}.
\begin{align}\label{eq:grasp_def}
g=(P,\phi, w).
\end{align}

\textbf{Transformation}  Grasp representation and policy generation take place in the image space and camera coordinate. The transformation ${}_{Image}^{Camera}T$ between the  image space and the camera space can be calculated by intrinsic parameters of the camera. We obtain transformation from the camera space to the robot base  ${}_{Camera}^{Robot}T$ by hand-eye calibration. 

The RGB image with the depth image aligned to it is $I = \mathbb{R}^{4\times H\times W}$ where $H$ is the image height and $W$ is the image width. We assume the intrinsic and extrinsic parameters of the camera and physics properties of the robot are known.
In the image space, the grasp representation can be rewritten as:
\begin{align}
{\widetilde g} = (u,v,\widetilde\phi,\widetilde\omega),
\end{align}
where $(u,v)$ is the position of the grasp in the image coordination. $\widetilde\phi$ and $\widetilde\omega$ correspond the $\phi$ and $w$ in Cartesian coordinates. We can get the grasp representation in the robot base coordinates following the Eq. \eqref{eq:convert2Cartesian}. 

\begin{align}\label{eq:convert2Cartesian}
g = {}_{Camera}^{Robot}T \times {}_{Image}^{Camera}T \times \widetilde{g}.
\end{align}

\subsection{Grasp labeling}\label{sub:grasping_label}
Most existing grasping methods assume that the center of the target bounding box is the best grasping center point and the length and angle of the short side of the bounding box represent the gripper's best opening width and orientation. This center-based assumption is used widely in 2D and 3D bounding box. 

\begin{figure}[htbp]
	\centering
	\subfigure[Object image]{
		\begin{minipage}[t]{0.25\linewidth}
			\centering
			\includegraphics[width=0.8in]{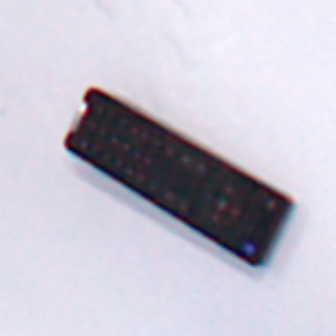}
		\end{minipage}
	}
	\subfigure[Mask image]{
		\begin{minipage}[t]{0.25\linewidth}
			\centering
			\includegraphics[width=0.8in]{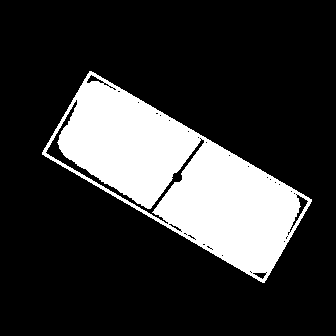}
		\end{minipage}
	}
	\subfigure[Mask image with the elliptical window]{
		\begin{minipage}[t]{0.25\linewidth}
			\centering
			\includegraphics[width=0.8in]{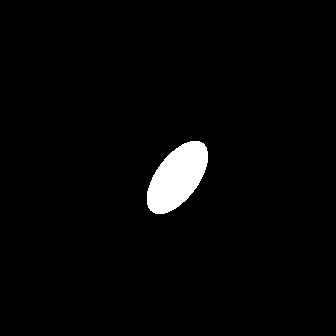}
		\end{minipage}
	}

	\centering
	\caption{\textbf{Grasp Labeling Process}}
	\label{Grasp_label}
\end{figure}

Unlike other robotic manipulation label method using rectangles and bounding boxes, we propose a novel labeling method for the common datasets that contain RGB, depth and object's mask images. The process can be shown in Fig. \ref{Grasp_label}. We also assume that the geometric center of the object is the best grasping center point as people usually grasp an object at the center of the object(geometric or gravity center). In addition, the shortest edge point-pair line passing the center is the best grasping instead of the bounding box's short-side direction. We use the minimum bounding box method to get a bounding box over the mask target image. This operation aims to obtain the geometric center of the target approximately. A shortest straight-line pass through two edge points and center point of the object is selected to get the gripper's opening width and orientation.
	
After obtaining the grasp representation ${\widetilde g}$, we need to label the positive pixel area as the ground- truth. Because we have only one grasp point label from the mask image, it is too sparse to train. We propose an elliptical window to label positive area as shown in Fig. \ref{Grasp_label}. The elliptical window uses grasp center point as center, the shortest straight-line as long axis and half lengths as short axis. We assume the area within the window has a high probability to grasp successfully and set it to 1 and also the areas corresponding to width and angle image are set to the width and angle values respectively.
            

\subsection{Problem formulation}  \label{objective}
RGB-D fusion grasping generation can be defined as a mapping $M: I \to G$. $I$ indicates the RGB-D image space and $G$ is the grasp representation space which can be redefined as 
\begin{align}
G = (Q, \widetilde{\Phi}, \widetilde{W}),
\end{align}
where $Q, \widetilde{\Phi}, \widetilde{W} \in \mathbb{R}^{ H\times W}$ is each pixel's probability, orientation and  opening width respectively. We get the position of grasp $(u, v)$ by selecting the pixel that has the maximum probability.

A mapping function $M$ can be defined as:
\begin{align}\label{mapping_func}
G = M(I).
\end{align}

Our goal is to find a robust function $M_\theta$ to fit $M$:
\begin{align}
\theta  = \mathop {\arg \min }\limits_\theta  \mathcal{L}\left( {G,{M_\theta(I) }} \right),
\end{align}
where $\mathcal{L}$ is the loss function between the ground-truth and $M_\theta$, $\theta$ is the parameter of function $M$. The optimal grasp $\widetilde g^* = \mathop {\max }\limits_Q G$ in the image space and grasp representation in robot based coordinates can be transferred via Eq. \eqref{eq:convert2Cartesian}.

\section{The Proposed Method}\label{sec:Proposed_method}
In this section, we will introduce a multimodal hierarchical encoder-decoder fully convolutional network to approximate the mapping function \eqref{mapping_func}. The dataset preprocess and the loss function definition are also given. A grasping metric is proposed to evaluate our method.

\subsection{Dataset generation and preprocessing}\label{sec:data_generation}
We use Cornell grasp detection dataset\cite{lenz2015deep} and YCB Object and Model Set\cite{calli2015benchmarking} to train our neural network. We extract an object mask image for the first one.
We generate ground-truth of grasping probability, gripper's opening width and orientation by mask image talked in Sec. \ref{sub:grasping_label}.

For the orientation, we choose a gripper orientation angle $\phi$ in the range of  $[-\frac{\pi}{2},\frac{\pi}{2}]$ and represent $\widetilde\phi$ as a vector $(\cos(2\widetilde\phi), \sin(2\widetilde\phi))$ on a unit circle of which value is a continuous distribution in $[-1, +1]$. \cite{hara2017designing} shows this processing is easy for the training. 
The input and output images are resized into $336 \times 336$. We scale the depth and RGB image value between 0 and 1 by min-max normalization. The depth image inpainting is used by OpenCV \cite{bradski2000opencv}.

\subsection{Network Design}
\begin{figure*}
	\centering
	\includegraphics[width=0.8\textwidth]{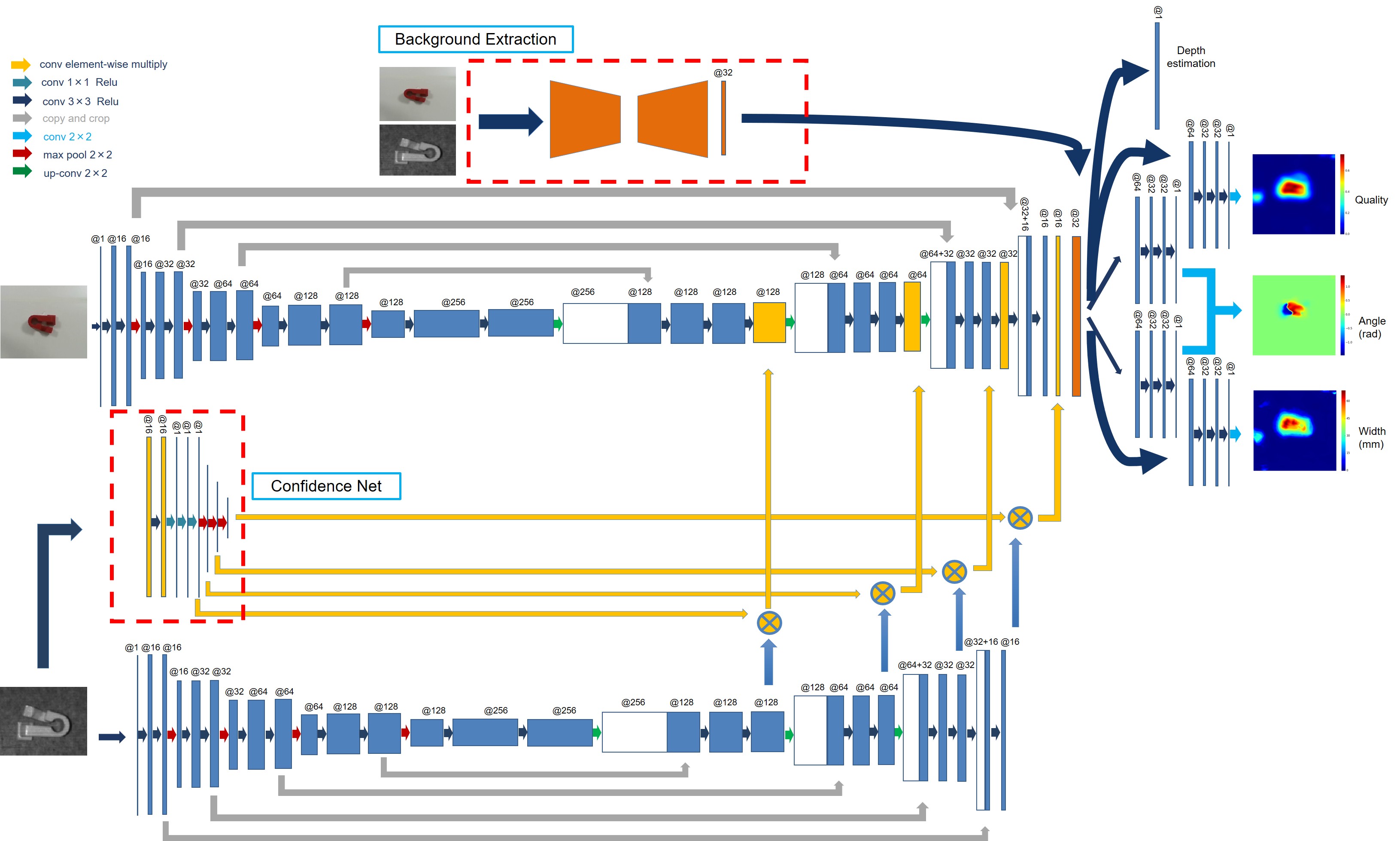}
	\caption{\textbf{UG-Net V3 Architecture} The network contains a two-input encoder-decoder FCN network, a background extraction module, and a confidence net. We proposed an aux depth estimation to compensate poor-quality depth image. The outputs are grasping probability, cosine value, sine value and gripper's opening width in pixel-wise respectively. } 
	\label{UG-Net_V3}
\end{figure*}

The architecture of our neural network is named as U-Grasping network V3(UG-Net V3). In the RGB branch and depth branch, we both adopt the structure like U-Net\cite{ronneberger2015u} as our back-bone illustrated in Fig. \ref{UG-Net_V3}. The last two layers of U-net is dropped and we reduce the channel of each layer to one-quarter of the original number except the input layer.

\textbf{Multimodal fusion} 
We fuse two modal features in the decoder part of the network by a confidence net(Confinet)\cite{zeng2019deep}. Four scale confidence maps are generated by confidence net via five convolutional layers and three max pooling layers to re-weight depth branch feature maps. At last, the re-weighted depth feature maps at four different scales are concatenated with color feature maps with the same resolution. The fusion module can be formulated as follows:
\begin{align}
FM\left( {\mathcal{F}_c^l,\mathcal{F}_d^l|{\mathcal{C}^l}} \right) = \mathcal{F}_c^l \oplus \left( {\mathcal{F}_d^l \odot {\mathcal{C}^l}} \right),
\end{align}
where ${\mathcal{F}_c^l}$, ${\mathcal{F}_d^l}$ are features from RGB and depth branches at scale $l$. ${\mathcal{C}^l}$ is the confidence map with only one channel at scale $l$. $\oplus$ is the concatenate operation and $\odot$ is the element-wise multiplication.

\textbf{Background extraction module(BEM)}
Most existing robotic grasping researches, excluding the ones using point cloud, set the camera heading vertically to the table surface, which adds a relatively strong constraint to the depth data. Since most laser-based RGB-D cameras obtain 3D data by a near-infrared (NIR) pulse illumination component no matter time-of-flight or structured light, the range of depth value is simply limitless comparing to RGB data range from 0 to 255 theoretically. The existing vertical observation constrains the depth value by the table surface which makes use of depth data more easily. However, this method lets the robotic arm grasp an object like humans difficultly to realize comparing with the method utilizing 3D point cloud data to predict 6D object pose. Therefore, to grasp without full observation and 3D prior knowledge, we propose an FCN mask network to extract the target object from the background. It is a six layer encoder-decoder neural network. We input RGB and depth image to a three-layer-encoder branch respectively and concatenate the last encoder features together passing to the decoder block. We set the output as the object mask by a linear convolutional layers. We pretrain this mask network using YCB Object and Model Set. and then concatenate the third decoder layer to the last fusion layer shown in the top of Fig. \ref{UG-Net_V3} (colored in orange) by dropping the last output layer. This part enhances our method grasping performance when our observation view is not vertical to the table surface even horizontal view.

\textbf{Grasping prediction}
We have introduced multimodal fusion and background extraction in the last two parts. In this part, we will illuminate the multi-task learning structure for the grasp representation. After concatenating the fusion features from RGB and depth branches and mask feature from background extraction, we add three convolutional layers and one output layers to predict grasping probability, cosine value, sine value and gripper's opening width respectively. In the physical system, the depth raw data from the RGB-D camera is filled with a lot of  missing points. We usually process the raw depth image by inpainting. For grasping task, we not only need to process visual  information to generate policy but also to obtain a real physical value not artificial to apply in the control system. Therefore, we propose an aux depth estimation output in the top right of Fig. \ref{UG-Net_V3}. Considering the last concatenating part fusing RGB features and re-weight depth features, we believe the RGB knowledge in the fusion features can compensate poor-quality depth image features implicitly and be also used to estimate missing points in the depth image for the control system.

\subsection{Proposed loss function and training}
\textbf{Loss function} For the background extraction module, we use mean square error loss Eq. \eqref{eq:loss_mask_net} function comparing to the predicted and ground-truth maps $M_\theta$ and $M^\star$. 
\begin{align}\label{eq:loss_mask_net}
\begin{array}{l}
	\mathcal{L}_{mask} = \frac{{1}}{n}{\left\| {{M_\theta } - M^\star} \right\|^2},
\end{array}
\end{align}

For the grasping prediction, we set the loss to:
\begin{align}
\mathcal{L} = \mathcal{L}_{depth} + \mathcal{L}_{grasp},
\end{align}
\begin{align}\label{eq:loss_depth}
          {{\rm{\mathcal{L}}}_{depth}}\left( {D, D^\star} \right) = \frac{\lambda_d}{n}\sum\limits_i {d_i^2} {\rm{ + }}\frac{\lambda_d}{n}\sum\limits_i {\left[ {{{\left( {{\nabla _x}{d_i}} \right)}^2}{\rm{ + }}{{\left( {{\nabla _y}{d_i}} \right)}^2}} \right]},
\end{align}
where $d=D - D^\star$, the sums are valid pixel $i$ and $n$ is the number of valid pixels. $D$ and $D^\star$ are the predicted and ground-truth of depth. And the first-order matching term encourages prediction of not only close-by value but also local structure\cite{eigen2015predicting}. 
\begin{align}\label{eq:loss_grasp}
\begin{array}{l}
\mathcal{L}_{grasp} = \frac{{{\lambda _q}}}{n}{\left\| {{Q_\theta } - Q^\star} \right\|^2} + \frac{{{\lambda _{\cos }}}}{n}{\left\| {{\cos(2\widetilde\Phi)_\theta } - \cos(2\widetilde\Phi)^\star} \right\|^2}\\
\\
+ \frac{{{\lambda _{\sin }}}}{n}{\left\| {{\sin(2\widetilde\Phi))_\theta } - \sin(2\widetilde\Phi))^\star} \right\|^2} + \frac{{{\lambda _w}}}{n}{\left\| {{\widetilde{W}_\theta } - \widetilde{W}^\star} \right\|^2}
\end{array},
\end{align}
where we train the gripper orientation using a vector on an unit circle talked in Sec. \ref{sec:data_generation}. We rewritten the vector $(\cos(2\widetilde\Phi), \sin(2\widetilde\Phi))$ and $\Phi$ can be computed following Eq. \eqref{eq:arctan}. $Q^\star$, $\cos(2\widetilde\Phi)^\star$, $\sin(2\widetilde\Phi))^\star$ and $\widetilde{W}^\star$ are corresponding ground-truth. $\lambda_d$, $\lambda _q$, $\lambda _{\cos}$, $\lambda _{\sin}$, $\lambda _w$ are the weight coefficients and we set all of default value as $1$.

\begin{align}\label{eq:arctan}
\widetilde\Phi_\theta = \frac{1}{2}arctan\frac{sin(2\widetilde\Phi_\theta)}{cos(2\widetilde\Phi_\theta)}.
\end{align}

\textbf{Training} We use 80\% of the dataset for training and 20\% of the dataset for evaluation. The activation function for all the layers except the output layers set as \textit{Relu} and output layers' set as \textit{linear}.  The training of our network proceeds into two steps. In the first step, we pretrain the background extraction module firstly using YCB Object and Model Set. In the second step, we load the pretrained parameters of module except the last output layers to our main network by concatenation and we set the last decoder of the mask network trainable and fix the parameters of other layers to fine-tune the module. The Cornell grasp detection dataset is used to train the main network. We use the Adam optimizer, set batch size as 4 and train 20 epochs for both two training steps.

%
\subsection{Grasping Controller}
We set up an open-loop grasping controller to realize grasping using our proposed method. We choose two ways to obtain grasping pose. The first one is to consistent with the observation viewpoint and the second one is obtain from object surface normal. The controller is presented in Algorithm \ref{algorithm:controller}.

\subsection{Grasping Metric}\label{grasp_metric}
\textbf{Robust grasp generation(RGG)} The variance of 200 times policy generation for a static observation. This evaluates the robustness of our method for the sampling noise.

\textbf{Success Rate(SR)} The percentage of successful grasps over all the grasp trails.

\textbf{Robust Grasp Rate(RGR)} The ratio of probabilities higher than 50\% of successful grasps over all the runnings.

\textbf{Planning Time(PT)} The time consumed between receiving the raw data and grasping policy generation output. 

\section{Experiment}
We design two physical experiments to validate the effectiveness of our method. The first one called grasping performance test, is to grasp an object with the camera view vertical to the table surface. The second one called 3D grasping test, is that we design a control algorithm to execute humanoid grasping operation.

\begin{figure}[htbp]
	\centering
	\subfigure[Adversarial set]{
	\begin{minipage}[t]{0.25\linewidth}
		\centering
		\includegraphics[width=0.9in]{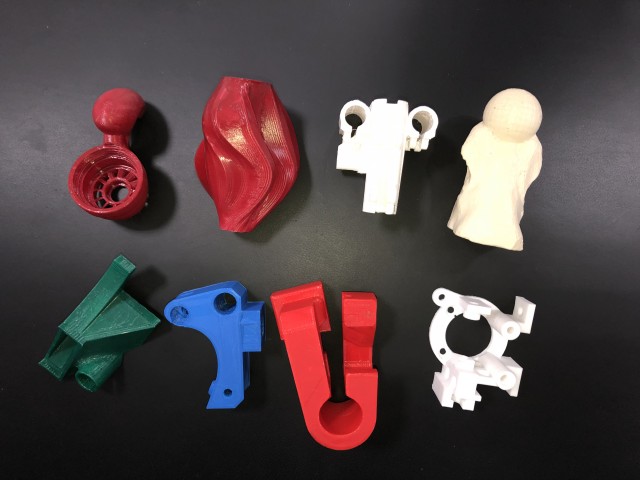}
		\label{test_exp1}
	\end{minipage}
	}
	\subfigure[Household set]{
	\begin{minipage}[t]{0.25\linewidth}
		\centering
		\includegraphics[width=0.9in]{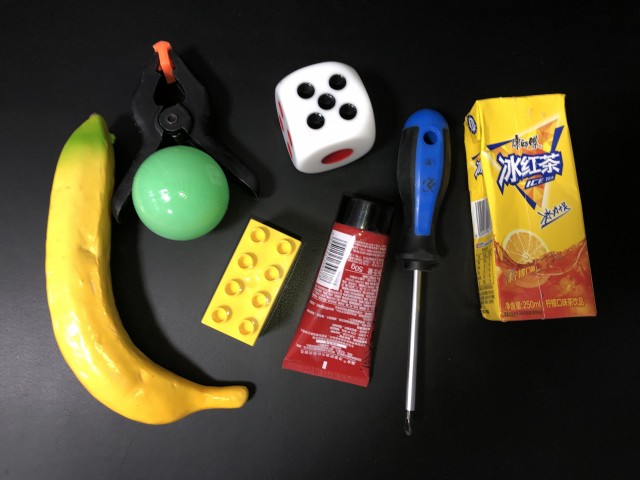}
		\label{test_exp2}
	\end{minipage}
	}
	\subfigure[Bottle set]{
	\begin{minipage}[t]{0.25\linewidth}
		\centering
		\includegraphics[width=0.9in]{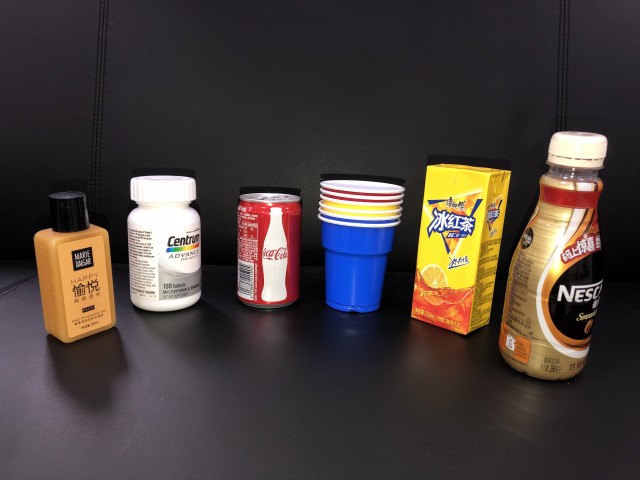}
		\label{test_exp3}
	\end{minipage}
	}
	\caption{\textbf{Test Objects}}
	\label{Test_Object}
\end{figure}

\begin{figure}
	\centering
	\subfigure[Grasping performance test]{
		\begin{minipage}[t]{0.4\linewidth}
			\centering
			\includegraphics[width=1.0in]{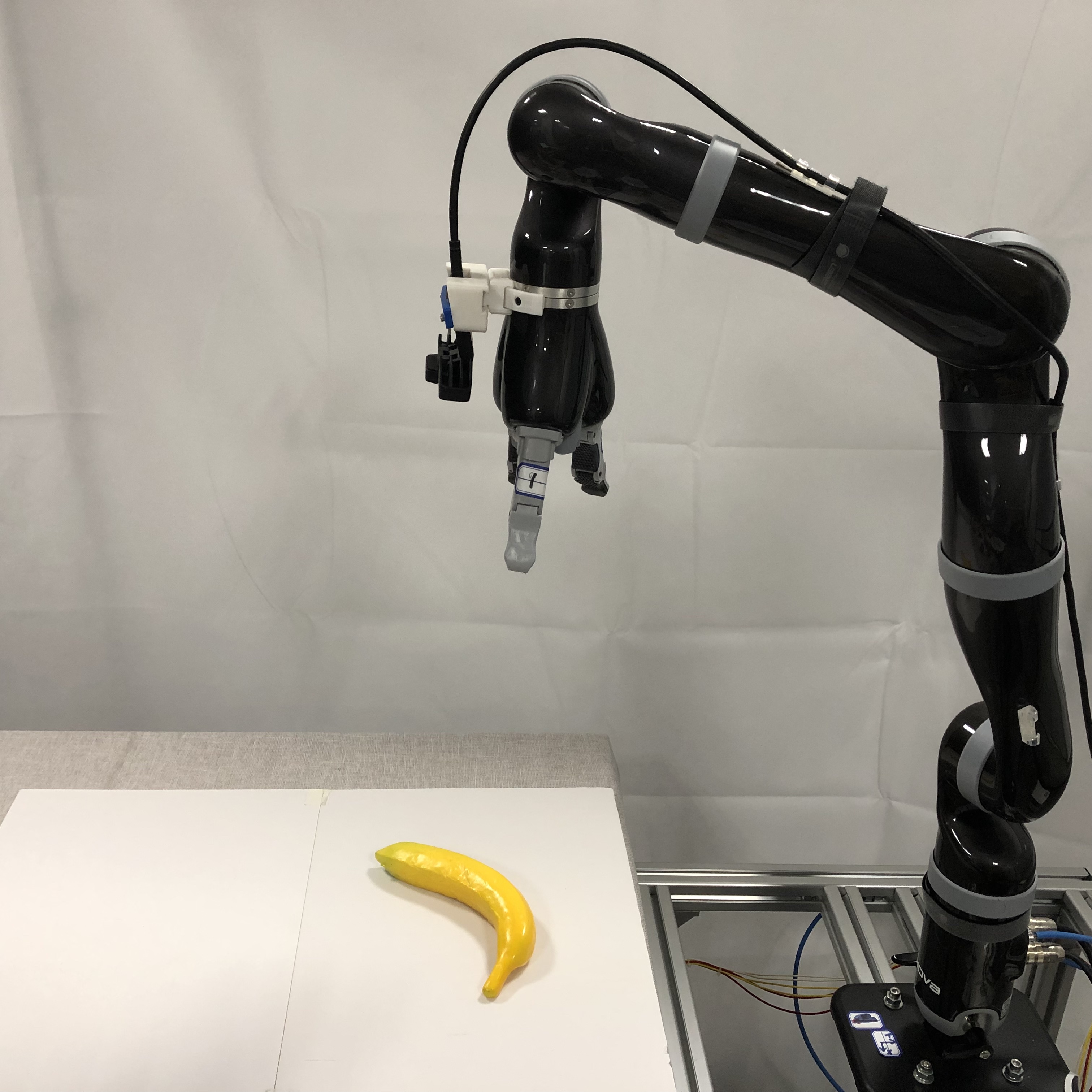}
			\label{set_up_exp1}
		\end{minipage}
	}
	\subfigure[3D Grasping test]{
		\begin{minipage}[t]{0.4\linewidth}
			\centering
			\includegraphics[width=1.0in]{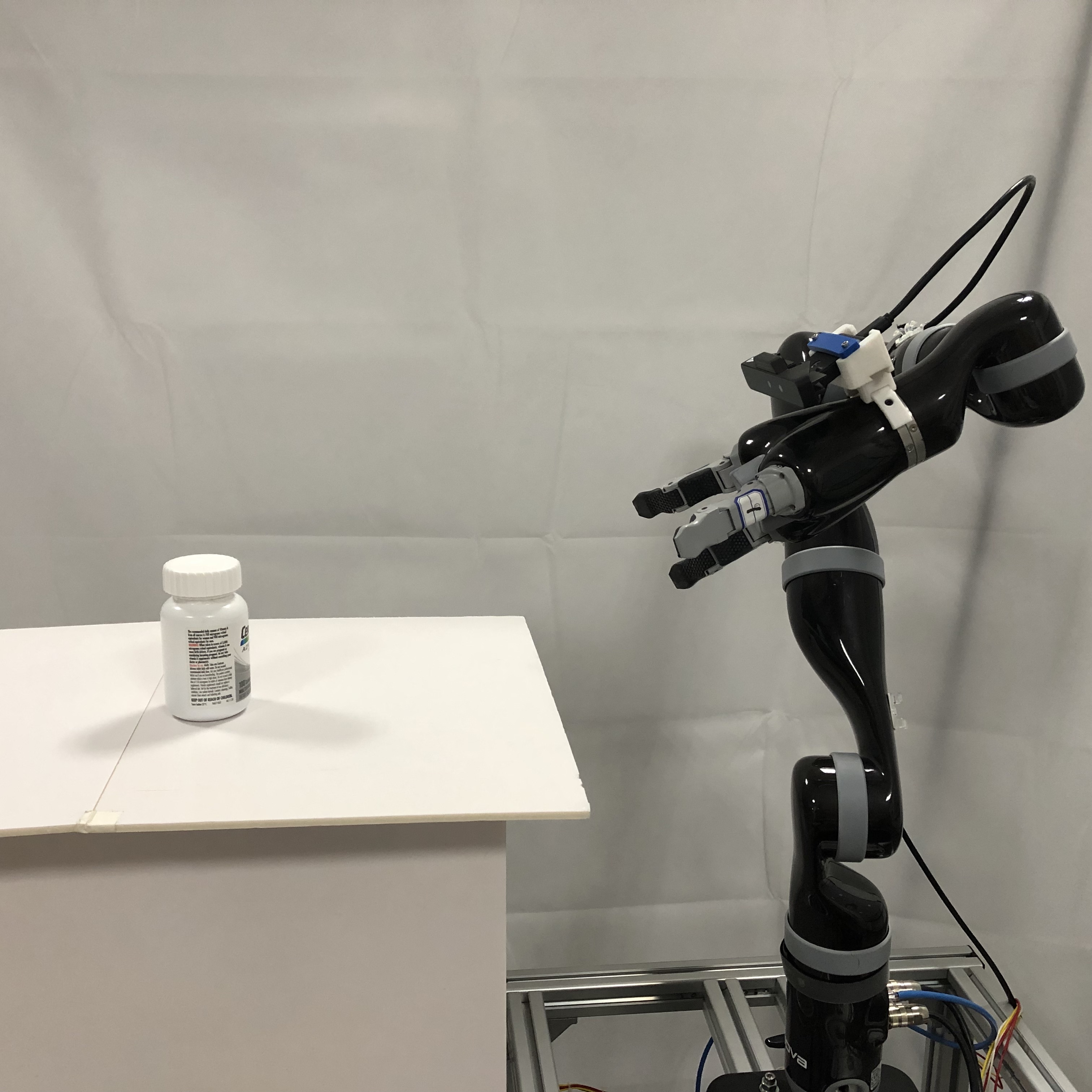}
			\label{set_up_exp2}
		\end{minipage}
	}
	\caption{\textbf{Robot Setup}}
	\label{set_up}
\end{figure}

We choose a standard test set including 3D-printed adversarial objects in Dex-Net 2.0 \cite{mahler2017dex} and household objects shown in Fig. \ref{Test_Object}. We also add a set of bottle-like object to validate the horizontal grasping. The final execution is taken place on a single arm Kinova Jaco 7DOF robot shown in Fig \ref{set_up}. We use an Intel RealSense SR300 RGB-D camera to get images mounted on the wrist of the robot. All the computation is finished on a PC running Ubuntu16.04 with an Intel Core i7-8700K CPU and two NVIDIA Geforce GTX 1080ti graphic cards. It is noticed that we use two cards to accelerate the training process and just use one card to run the trained network.

\subsection{Assumption}\label{sec:assume}
In our work, we assume that the intrinsic and extrinsic parameters of the camera are known. The coordinates of RGB and depth image are aligned and time stamp is synchronized. For the 3D grasping test, since we do not estimate the 6D pose of the object directly, we assume that the best grasping pose is approximately equal to the best camera viewpoint pose considering that the camera is fixed on the end-effector close to the end of gripper.

\subsection{Grasping Performance Test}\label{sub_vertical_table}
We make an analysis and comparison between the state of art methods in the metrics proposed in Sec. \ref{grasp_metric}. We also perform an ablation study to evaluate each module in our network structure.

We set up our robot as shown in Fig. \ref{set_up_exp1}. The observation height is 55 centimeters away from the table surface. We set the observation pose vertical to the table surface approximately which is the same as existing work. We try to grasp each object 10 times to record the grasping results shown in TABLE \ref{baseline_comparision}.

In the experiment, we set the network structure only containing RGB and depth encoder-decoder branch without the confidence net and background extraction module as our \textit{baseline}. All the network structures achieve a comparative high grasping success rate in the household set because some kinds of objects appeared in the training dataset. The network can better learn the geometric and textural features of these objects. In contrast, the result of adversarial sets show a little poor, but the network still achieves high success rate and robust grasp rate when we adopt the additional modules. We find that the baseline achieves 80\% performance in both success rate and robust grasp rate which indicates the advantage of our fusion network structure. The background extraction module splits the background and the target object to improve the grasping policy generation especially in the complex scenario. The top of Fig .\ref{module_compare} shows that Baseline+BEM improves the grasping precision comparing to only-depth method. The confidence net is used to quantify the uncertainty of depth image with sampling noise or missing points. While the confidence net decreases the robust grasp rate, it enhances our method’s robustness for the different scenarios which will be talked in Sec. \ref{sub_3d_test}. For the RGG, we find that our method can maintain it stably within low value. 

Apart from the ablation study, another comparison with some state-of-the-art approaches is shown in TABLE \ref{comparision_other_work}. Considering the test set used, we only compare adversarial set results with all mentioned work. We can find our method has a desiring balance between real-time and performance. Furthermore, in our work, the aux depth estimation can avoid the failure of grasping due to the depth points missing to some degree. For the real-time, GQ-CNN\cite{mahler2017dex} and GG-CNN\cite{morrison2018closing} use GTX 1080 and GTX 1070 GPU respectively while we adopt one GTX1080ti GPU to run grasp prediction. We can maintain over 25HZ to keep the real-time for the robot control.

\subsection{3D Grasping Test}\label{sub_3d_test}
\begin{figure}
	\centering
	\subfigure[Pitch angle $90^\circ$]{
		\begin{minipage}[t]{0.25\linewidth}
			\centering
			\includegraphics[width=0.8in]{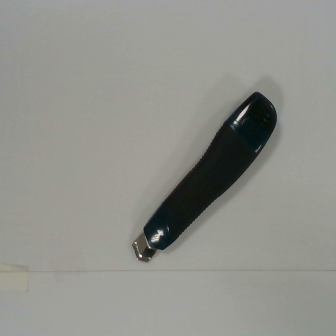}
			\label{module_compare1_1}
		\end{minipage}
	}
	\subfigure[Depth-only]{
		\begin{minipage}[t]{0.25\linewidth}
			\centering
			\includegraphics[width=0.8in]{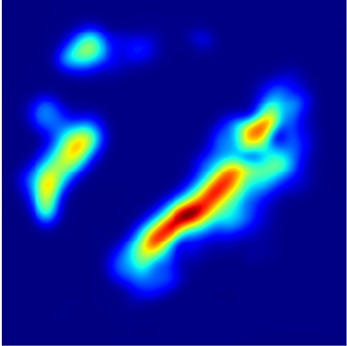}
			\label{module_compare1_2}
		\end{minipage}
	}
	\subfigure[Baseline+BEM]{
		\begin{minipage}[t]{0.25\linewidth}
			\centering
			\includegraphics[width=0.8in]{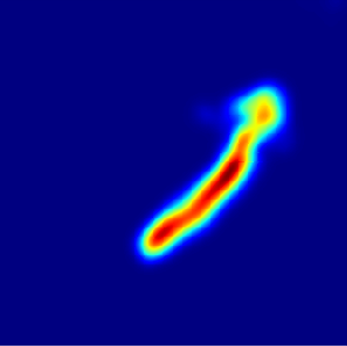}
			\label{module_compare1_3}
		\end{minipage}
	}
	
	\subfigure[Pitch angle $0^\circ$]{
		\begin{minipage}[t]{0.25\linewidth}
			\centering
			\includegraphics[width=0.8in]{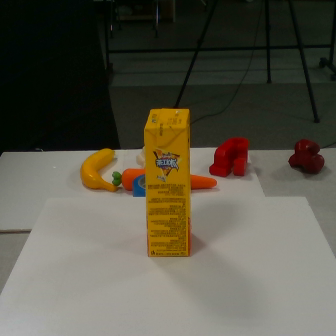}
			\label{module_compare2_1}
		\end{minipage}
	}
	\subfigure[Depth-only]{
		\begin{minipage}[t]{0.25\linewidth}
			\centering
			\includegraphics[width=0.8in]{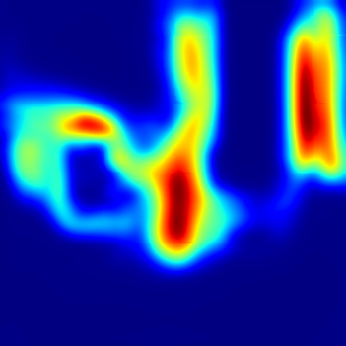}
			\label{module_compare2_2}
		\end{minipage}
	}
	\subfigure[Confinet used]{
		\begin{minipage}[t]{0.25\linewidth}
			\centering
			\includegraphics[width=0.8in]{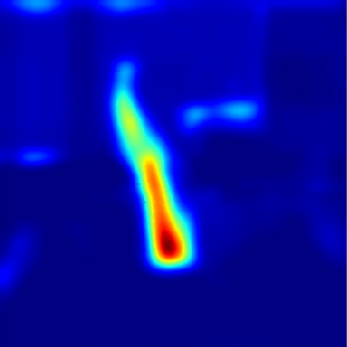}
			\label{module_compare2_3}
		\end{minipage}
	}
	\caption{\textbf{Result Visualization in Heatmap}}
	\label{module_compare}
\end{figure}

We use the eye-in-hand camera system that the camera is fixed very close to the end-effector. we assume there is a consistency between observation viewpoint and grasping pose and the best observation viewpoint is approximately equal to the best grasping pose. We propose a test set including six objects shown in Fig. \ref{test_exp3} observed at $0^\circ, 22.5^\circ, 45^\circ$ horizontal angles(pitch angles) which are same as BigBIRD\cite{singh2014bigbird} setup shown in Fig. \ref{set_up_exp2}. We only choose the three low pitch angle setups since the other two approximately observe vertical to the table surface. The table surface guarantees the quality of depth image with less missing points while observation with low pitch angle samples more invalidated data because of the perception range of the sensor. We have 10 trials for each object which is rotated along the gravity axis with a random angle and a grasp success rate is given in TABLE \ref{baseline_comparision}. From the result, we find different observation viewpoint has different grasping quality in the partial observation scenario. The background extraction module and confidence net improve grasping performance obviously because of the uncertainty quantification of depth image and detection of complex background. The bottom of Fig. \ref{module_compare} shows that the Baseline+Confinet can quantify the missing points effectively in the depth image which leads to false prediction.

We also try to design a grasping based on the surface normal estimation to replace the grasping pose from the observation viewpoint. we adopt the surface normal estimation\cite{qi2018geonet} which takes about 470ms for each policy generation and set the normal vector of target grasping point align to the z-axis of camera coordinate within the shortest arc.   

%

\begin{table}[]
	\centering
	{\small
	\begin{tabular}{ccccc}
		\hline
		\textbf{\begin{tabular}[c]{@{}c@{}}Structure \end{tabular}} & \textbf{Baseline}    & \textbf{\begin{tabular}[c]{@{}c@{}}Baseline\\ Confinet\end{tabular}} & \textbf{\begin{tabular}[c]{@{}c@{}}Baseline\\ BEM\end{tabular}} & \textbf{\begin{tabular}[c]{@{}c@{}}Baseline\\ Confinet\\ BEM\end{tabular}} \\ \hline
		\multicolumn{1}{l|}{\textbf{SR(adver.)}}            & 79$\pm$2\%               & 73$\pm$2\%                                                                    & \textbf{95+1\%}                                                                 & \textbf{89+1\%}                                                                                  \\ \hline
		\multicolumn{1}{l|}{\textbf{RGR(adver.)}}       & 80$\pm$1\%               & 31$\pm$2\%                                                                    & 97$\pm$1\%                                                                          & 79$\pm$2\%                                                                                           \\ \hline
		\multicolumn{1}{l|}{\textbf{RGG(adver.)}}       & 
		5.05     & 
		1.32                                                          & \textbf{1.30}                                                                         & 1.53                                                                                          \\ \hline
		\multicolumn{1}{l|}{\textbf{SR(house.)}}              & 96$\pm$1\%               & 97$\pm$2\%                                                                    & 99$\pm$1\%                                                                         & 93$\pm$2\%                                                                                           \\ \hline
		\multicolumn{1}{l|}{\textbf{RGR(house.)}}         & 97$\pm$1\%               & 87$\pm$1\%                                                                    & 98$\pm$2\%                                                                          & 75$\pm$2\%                                                                                           \\ \hline
		\multicolumn{1}{l|}{\textbf{RGG(house.)}}         & 
		\textbf{1.21} & 
		3.24                                                      & 3.22                                                            & 
		3.60 
	    \\ \hline
	  	\multicolumn{1}{l|}{\textbf{Pitch angle: $0^\circ$}}    & 80$\pm$1\%            & 64$\pm$1\%                                                                     & \textbf{86$\pm$2\%}                                                                           & 81$\pm$1\%                                                                                            \\ \hline
	  	\multicolumn{1}{l|}{\textbf{Pitch angle: $22.5^\circ$}} & 80$\pm$1\%            & 84$\pm$1\%                                                                     & 83$\pm$2\%                                                                           & \textbf{93$\pm$1\% }                                                                                           \\ \hline
	 	\multicolumn{1}{l|}{\textbf{Pitch angle: $45^\circ$}}   & 67$\pm$1\%            & \textbf{90$\pm$1\% }                                                                    & \textbf{90$\pm$1\%}                                                                           & 85$\pm$1\%                                                                   \\ \hline
	\end{tabular}}
	\caption{Results of the ablation study in Sec. \ref{sub_vertical_table} on the test set. Different pitch angle grasping is also given mentioned in Sec. \ref{sub_3d_test}}
	\label{baseline_comparision}
\end{table}

\begin{table}[]
	\centering	
	{\scriptsize
	\begin{tabular}{lcccc}
		\hline
		& \textbf{ GQ-CNN\cite{mahler2017dex}} & \textbf{ GG-CNN\cite{morrison2018closing}} & 
		\textbf{ \begin{tabular}[c]{@{}c@{}}Our\\ baseline\end{tabular}} & 
		\multicolumn{1}{l}{\textbf{ \begin{tabular}[c]{@{}c@{}}Our best\end{tabular}}} \\ \hline
		\multicolumn{1}{l|}{\textbf{Success Rate}}        & 80\%            & 84$\pm$8\%          & 79$\pm$2\%                & 95$\pm$1\%                                            \\ \hline
		\multicolumn{1}{l|}{\textbf{\begin{tabular}[c]{@{}l@{}}Robust \\Grasp Rate\end{tabular}}}   & 43\%            &                 & 80$\pm$1\%                & 97$\pm$1\%                                            \\ \hline
		\multicolumn{1}{l|}{\textbf{\begin{tabular}[c]{@{}l@{}}Planning \\Time(ms)\end{tabular}}}   & 800             & 19              & 28                    & 32                                                \\ \hline
		\multicolumn{1}{l|}{\textbf{\begin{tabular}[c]{@{}l@{}}Model Size\\(Approx.)\end{tabular}}} & 75MB            & 0.5MB           & 19.1MB                  & 19.6MB                                              \\ \hline
	\end{tabular}}
	\caption{Comparison with other existing work from the metrics proposed in Sec. \ref{grasp_metric} }
	\label{comparision_other_work}
\end{table}


\begin{algorithm}[t!]
\KwIn  {RGB image $I_{RGB}$, depth data $I_{depth}$.}
\KwOut{ grasping $g$ and gripper $q$.}
\BlankLine
\While{Running}{
	\BlankLine
	Preprocess $I_{RGB}$ and $I_{depth}$.
	Predict $G$ from network $M_\theta$.
	Obtain the maximum probability position $\widetilde{g}$ in pixel space.
	\BlankLine
	\uIf{Target depth is missing}{
		Use estimated depth value instead.
	}
	Convert $\widetilde{g}$ to $g$ according to  Eq. \eqref{eq:convert2Cartesian}.
	\BlankLine
	\uIf{Surface normal is estimated}{
		Convert target point $\widetilde{g}$ normal vector with $(0, 0, 1)$ into quaternion $q$.
	}
	\Else{
		Obtain current pose of camera view point $q$.
	}
	Approach target object with position $g$ and pose $q$ and close the gripper.
}
\caption{Hierarchical RGB-D Fused Robotic Grasp Planning}
\label{algorithm:controller}
\end{algorithm}

\section{Conclusion} \label{sec:conclusion}
In this paper, we propose a novel multimodal hierarchical neural network to realize robotic grasping in an unstructured environment. Our method only adopts partial observation obtained by an eye-in-hand robotic arm system. Extensive experiment results including grasping pose vertical to the table surface and 3D humanoid grasping are given to validate effectiveness and performance. Our grasping method achieves over 90\% success rate on a standard test set and is competitive to other state-of-art methods.  


Nevertheless, the proposed approach can still be improved in the following aspects. In our work, we assume the grasping pose keeps consistent with the camera pose, while the best grasping pose is approximately equal to the best camera viewpoint. The gripper's grasping pose is obtained from the camera's observation pose, and it remains open on how to get the best observation. Exploration of the observation viewpoint will be studied in future research.

\bibliography{Feasibility}
\bibliographystyle{ieeetr}

\end{document}